%% file: 00_main.tex
\pdfoutput=1

\documentclass[11pt]{article}

\usepackage[final]{acl}

\input{preamble}

\title{\toolname: An Efficient Approach \\ for AI Safety Moderation Across Languages and Modalities}

\author{ 
\textbf{Sahil Verma}\affmark{1} \quad 
\textbf{Keegan Hines}\affmark{2} \quad 
\textbf{Jeff Bilmes}\affmark{1} \quad 
\textbf{Charlotte Siska}\affmark{2} \quad 
\\
\textbf{Luke Zettlemoyer}\affmark{1} \quad
\textbf{Hila Gonen}\affmark{1} \quad 
\textbf{Chandan Singh}\affmark{2} \quad 
\\
\affmark{1}University of Washington \quad
\affmark{2}Microsoft \quad
}

\begin{document}
\maketitle

\begin{abstract}
    \input{000_abstract}
\end{abstract}

\input{01_intro}

\input{02_methodology}

\input{03_experimental_setup}

\input{04_results}

\input{05_analysis-multilingual}

\input{mini_related_work}

\input{conclusions}
\input{limitations}

\FloatBarrier

{
\small
\bibliography{refs}
}

\clearpage
\appendix
\input{appendix}

\end{document}

%% file: preamble.tex
\usepackage{times}
\usepackage{latexsym}

\usepackage[T1]{fontenc}
\usepackage[utf8]{inputenc}
\usepackage{inconsolata}
\usepackage{graphicx}

\usepackage[utf8]{inputenc} 
\usepackage[T1]{fontenc}    
\usepackage{hyperref}       
\usepackage{url}            
\usepackage{booktabs}       
\usepackage{amsfonts}       
\usepackage{nicefrac}       
\usepackage{microtype}      
\usepackage{xcolor}         

\usepackage[accsupp]{axessibility}  

\usepackage{multicol}
\usepackage{etoolbox} 
\usepackage{xspace}
\usepackage{enumitem}

\usepackage{orcidlink}
\usepackage{acronym}
\usepackage[capitalize]{cleveref}
\usepackage{wrapfig}
\usepackage{adjustbox}
\usepackage{caption}
\usepackage{multirow}
\usepackage{breqn}
\usepackage{makecell} 
\usepackage{subcaption}
\usepackage{listings}
\usepackage{hyperref}       
\usepackage{tikz}
\usetikzlibrary{positioning}
\usetikzlibrary{calc}    
\usetikzlibrary{tikzmark}
\newcommand{\toolname}{\textsc{OmniGuard}\xspace}
\newcommand{\alignmentscorelong}{Universality Score\xspace}
\newcommand{\alignmentscore}{U-Score\xspace}
\newcommand{\affmark}[1]{\textsuperscript{#1}}

\definecolor{bblue}{rgb}{0.0, 0.6, 0.8}

\usepackage{placeins}

\usepackage{tabularx}

\newcommand{\setsectionlabel}[1]{%
  \setcounter{table}{0}%
  \setcounter{figure}{0}%
  \renewcommand{\thetable}{#1\arabic{table}}%
  \renewcommand{\thefigure}{#1\arabic{figure}}%
  \renewcommand{\theHtable}{#1\arabic{table}}%
  \renewcommand{\theHfigure}{#1\arabic{figure}}%
}

%% file: 000_abstract.tex
The emerging capabilities of large language models (LLMs) have sparked concerns about their immediate potential for harmful misuse.
The core approach to mitigate these concerns is the detection of harmful queries to the model. 
Current detection approaches are fallible, and are particularly susceptible to attacks that exploit mismatched generalization of model capabilities (e.g., prompts in low-resource languages or prompts provided in non-text modalities such as image and audio).
To tackle this challenge, we propose \toolname, an approach for detecting harmful prompts across languages and modalities. 
Our approach (i) identifies internal representations of an LLM/MLLM that are aligned across languages or modalities and then (ii) uses them to build a language-agnostic or modality-agnostic classifier for detecting harmful prompts. 
\toolname improves harmful prompt classification accuracy by 
11.57\% over the strongest baseline in a multilingual setting, 
by 20.44\% for image-based prompts,
and sets a new SOTA for audio-based prompts.
By repurposing embeddings computed during generation, \toolname is also very efficient (\textbf{$\approx\!120 \times$} faster than the next fastest baseline).
Code and data are available at \url{https://github.com/vsahil/OmniGuard}.



%% file: 01_intro.tex
\section{Introduction}
\label{sec:intro}

\begin{figure}
    \centering
    \includegraphics[width=\columnwidth]{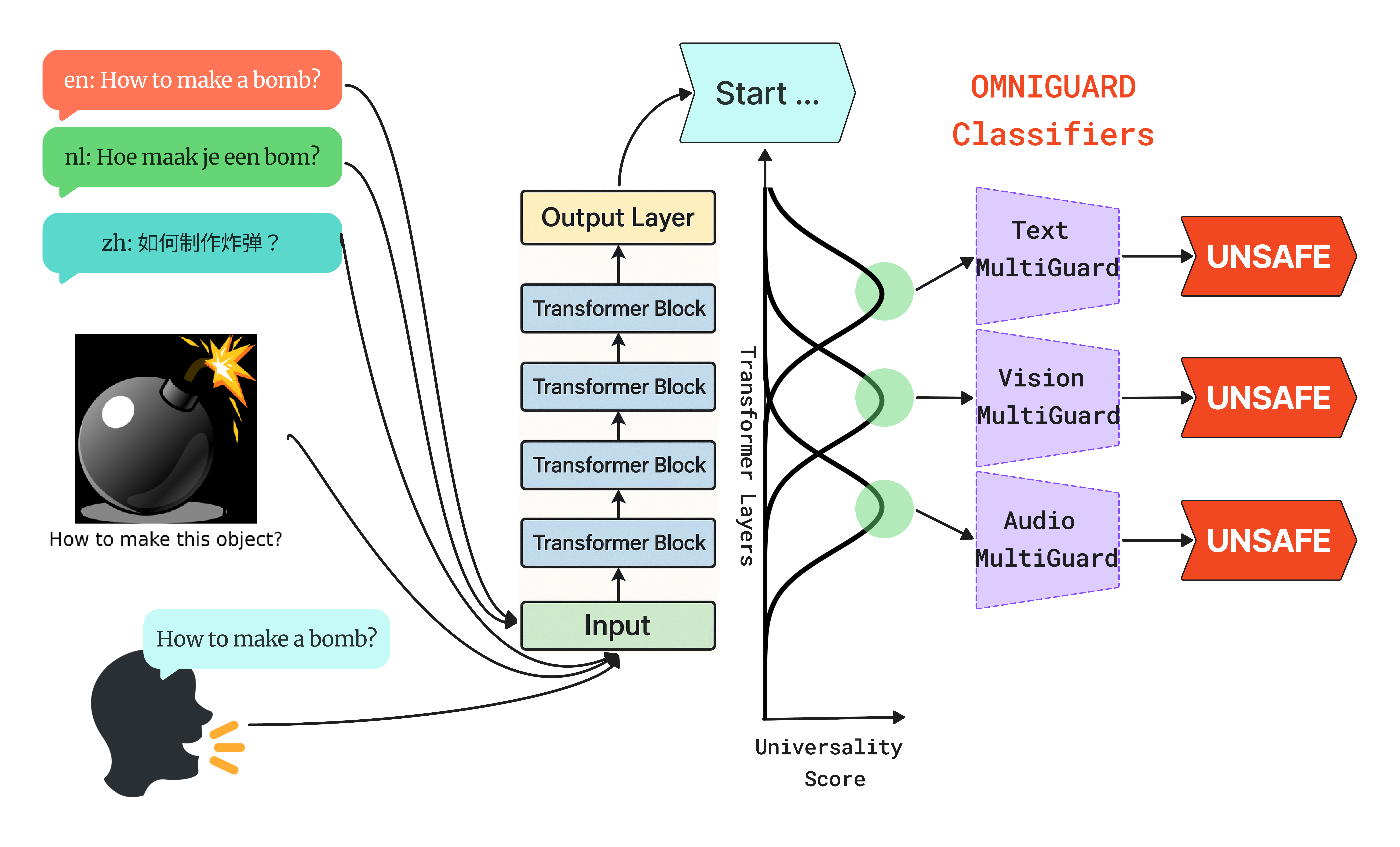}
    \caption{\toolname builds a harmfulness classifier that operates on internal representations of an LLM (or MLLM).
    \toolname uses a custom metric (\alignmentscore) to identify representations that generalize across languages and modalities.
    At inference time, \toolname re-uses the embeddings from the LLM/MLLM being used for generation, and thereby completely avoids the overhead of passing the inputs through a separate guard model for safety moderation. 
    }
    \label{fig:main-figure}
    \vspace{-1.5em}
\end{figure}

The rapid rise of capabilities in large language models (LLMs) has created an urgent need for safeguards to prevent their immediate harmful misuse as they are deployed to human users en masse~\cite{bommasani2022-opportunities-risks}.
Moreover, these safeguards are critical for defending against future potential harms from LLMs~\cite{bengio2024managing}. Standard safeguard approaches broadly include approaches such as safety training using reinforcement learning from human feedback~\citep{instruct-gpt,leike2018-scalable-agent-alignment} or using pre-trained guard models that classify the safety of an input prompt
~\citep{openai2025moderation,inan2023-llama-guardrails,Han2024-wildguardmix-dataset}. 

With these safeguards in place, harmful prompts in high-resource languages, e.g., English, are successfully detected. However, harmful prompts in low-resource languages can often bypass these safeguards~\citep{deng2024-multilingual-attack-firstpaper,yong2024-low-resource-languages-jailbreak-gpt4,dan-roth-paper-guard-llms-on-multilingual-data}, i.e.,  \textit{jailbreaking} the LLM. 
Modern LLMs are vulnerable to attacks not only from low-resource natural languages, but also from artificial \textit{cipher languages}, e.g., base64 or caesar encoding of English prompts~\citep{jacob-steinhardt-why-jailbroken,yuan2024-gpt-cipher-attack}. 
This phenomenon also extends beyond text to jailbreaking multimodal LLMs (MLLMs) using modalities such as images~\citep{figstep-dataset-image-jailbreak,mm-safetybench-image-jailbreak}
or audio~\citep{achilles-dataset-audio-jailbreaks}. 

\citet{jacob-steinhardt-why-jailbroken} argue that these attacks are successful due to \textit{mismatched generalization}, a scenario in which the model's safety training does not generalize to other settings, but general performance does. 
This may happen because pretraining data often includes more diverse data than that available for safety finetuning \citep{ghosh2024aegis2}.  
In this work, we defend against attacks that exploit the mismatched generalization of the safety training of LLMs and MLLMs. 
Specifically, we defend against attacks that utilize low-resource languages, both natural and cipher languages, as well as attacks employing other modalities, such as images and audio.

We introduce \toolname, an approach that builds a classifier using the internal representations of a model.
These representations are extracted from specific layers that produce representations that are universally similar across multiple languages and across multiple modalities. 
\toolname's classifier trained on such representations, is able to accurately detect harmful inputs across 73 languages, with an average of 86.22\% accuracy across 53 natural languages and an average of 73.06\% accuracy across 20 cipher languages. 
\toolname can also detect harmful inputs provided as images with 88.31\% and as audio with 93.09\% accuracy respectively. 

In contrast to popular guard models such as LlamaGuard~\citep{inan2023-llama-guardrails}, AegisGuard~\citep{ghosh2024-aegis-safety-guard}, or WildGuard~\citep{Han2024-wildguardmix-dataset}, \toolname does not require training a separate LLM specifically to detect harmfulness. 
By building a classifier that uses the internal representations of the main LLM or MLLM, \toolname avoids the overhead of passing the prompt through a separate guard model, making it very efficient.

In summary, our contributions are the following:
(1) We propose \toolname, an approach for detecting harmful prompts,
(2) we show that \toolname accurately detects harmfulness across multiple languages and multiple modalities,
(3) we show that \toolname is very sample-efficient during training,
and (4)
we show that \toolname is highly efficient at inference time.

%% file: 02_methodology.tex
\section{Methodology}
\label{sec:method}

\toolname seeks to robustly detect harmful prompts, regardless of their language or modality. 
We first leverage the tendency of LLMs and MLLMs to create universal representations that are similar across languages~\citep{wendler2024-convert-things-to-english,zhao2024-convert-things-to-english} and across modalities~\citep{wu2024-semantic-hub-hypothesis,zhuang2024vector} in \cref{sec:select-language-agnostic-part}, and then use them to train harmfulness classifiers that robustly detect harmful inputs in \cref{subsec:methods_classifier}.

\subsection{Finding language-agnostic representations in an LLM}
\label{sec:select-language-agnostic-part}

The first step of \toolname 
searches for internal representations of an LLM that are universally shared across languages. 
We prompt an LLM with English sentences and their translations to other languages, and extract their representations at different layers.\footnote{The representation of a prompt is computed by averaging the representation over each token in the prompt.} 
For language-agnostic representations, we expect the similarity between the representations of English sentences and the representations of their translations to be similar, and we expect this similarity to be higher than the similarity between representations of two sentences that are not translations of each other (a random pair of sentences). 
We concretize this notion by defining the \textit{\alignmentscorelong} (\textit{\alignmentscore}, Eq. \ref{eq:cross-lingual-score}), which is the difference between the average cosine similarities of pairs of sentences that are translations of each other and pairs of sentences that are not.
\begin{equation}
\label{eq:cross-lingual-score}
\begin{aligned}
    \text{\textit{\alignmentscore}} &:= \\
    & \frac{1}{N} \sum_{\substack{i \in [N]}} \text{CosSim} \left( \text{Emb}(e_i), \text{Emb}(l_i) \right) \\
    - \frac{1}{N(N-1)} & \sum_{\substack{i, j \in [N] \\ i \neq j}} \text{CosSim} \left( \text{Emb}(e_i), \text{Emb}(l_j) \right)
\end{aligned}
\end{equation}
where $e_i$ and $l_i$ are sentences in English and their translations to another language.

\begin{figure}
    \centering
    \begin{tikzpicture}
        \node[anchor=center] (img) at (0,0) {\includegraphics[width=\columnwidth]{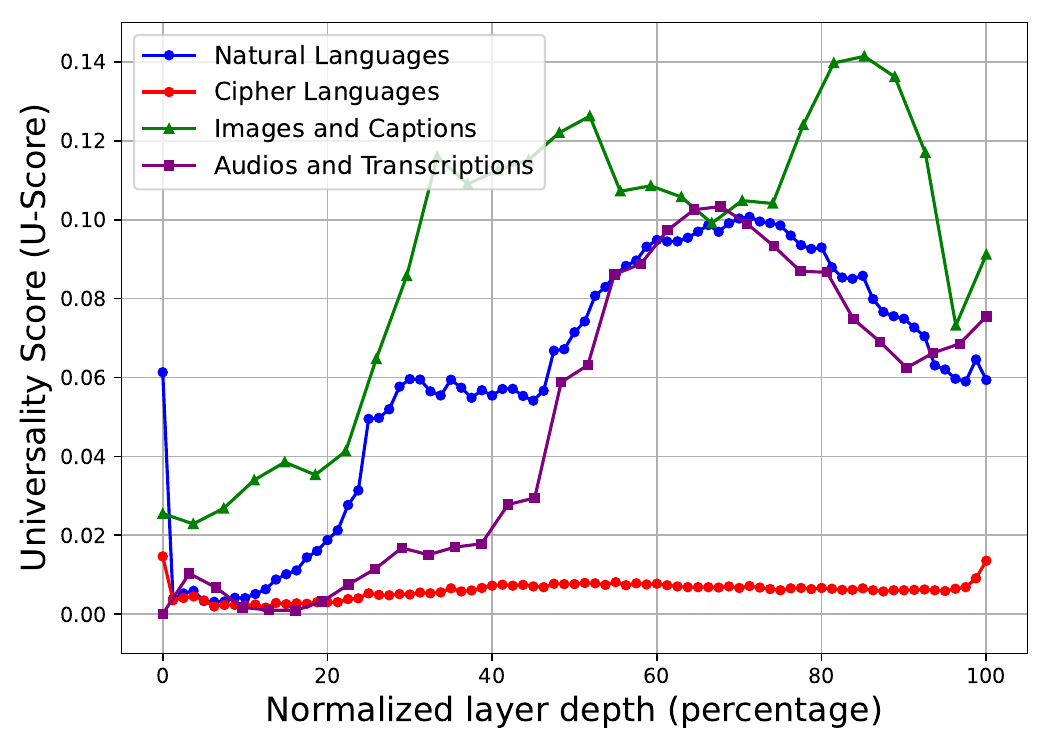}};
        \useasboundingbox (img.south west) rectangle (img.north east);
    \end{tikzpicture}
    
    \caption{The \textit{\alignmentscore} across different layers for different modalities. 
    (A) Different layers of the model Llama3.3-70B-Instruct for different languages.
    (B) The \textit{Cross-Model Alignment Score} at different layers of the model (Molmo-7B) for similarity between images and captions. The highest values are obtained with at layers 21-25, indicating better alignment between images and their text captions at these layers. 
    (C) The \textit{Cross-Model Alignment Score} at different layers of the model (Llama-Omni 8B) for similarity between audios and transcriptions. The highest values are obtained with at layers 20-23, indicating better alignment between audios and their text transcriptions at these layers. 
    }
    
    \label{fig:difference-cosine-sim-aligned-shuffled}
    \vspace{-1em}
\end{figure}

This procedure can be generalized to new different modalities rather than different languages by changing which embeddings are being used.
For example, to determine if internal representations of an MLLM are aligned across modalities,
we replace embeddings for a translated piece of text with embeddings from a different modality (e.g. a text caption and its corresponding image, or a text transcription and its corresponding audio clip).
See experimental details in \cref{sec:setup-multilingual}. 

\subsection{Fitting a harmfulness classifier}
\label{subsec:methods_classifier}
After selecting the layer that maximizes the \alignmentscore{}, we extract embeddings from that layer and use them as inputs to fit a lightweight, supervised classifier that predicts harmfulness.
In our experiments, the classifier is a multilayer perceptron with 2 hidden layers (with hidden sizes 512 and 256).
At inference time, when a prompt is passed to a model for generation,
\toolname applies this classifier to the embeddings generated by the model, 
incurring minimal overhead at inference time for safety classification.
Note, however, that this approach only applies to open-source models, for which \toolname can build a classifier by obtaining embeddings.
During training, only the lightweight classifier's parameters are learned (the original model is never modified), making the training process data-efficient and inexpensive.

%% file: 03_experimental_setup.tex
\section{Experimental Setup}
\label{sec:setup-multilingual}

\Cref{tab:dataset_details} and \Cref{tab:model_details} give details on all the models and datasets for this section.

\input{tables/datasets-used}

\subsection{Selecting universal layers via the \alignmentscore}
\label{subsec:layer_selection}
\paragraph{Selecting language-agnostic layers}
To select language-agnostic layers, we use a dataset of translated sentence pairs spanning various languages.
Specifically, we use sentences in 53 natural languages from the Flores200 dataset
and additionally translate the sentences into 20 cipher languages (using encodings such as Caesar shifts, base64, hexadecimal);
see a full list in \cref{sec:languages-used}.
We extract embeddings from each layer of Llama3.3-70B-Instruct for the sentences in all 73 languages
and use them to compute the \alignmentscore (averaged over languages). 
\cref{fig:difference-cosine-sim-aligned-shuffled} shows the \alignmentscore as a function of layer depth.
For natural languages (blue curve), the \alignmentscore
peaks in the middle layers of the model, with the highest values in layer 57 (out of 81 layers).  
For cipher languages (red curve), the \alignmentscore is much lower than for natural languages, suggesting the model fails to represent semantic similarity in these languages (see analysis in \cref{sec:analysis}).

\paragraph{Selecting modality-agnostic layers}
To select layers aligned between images and captions, we use the MM-Vet v2 dataset, a popular dataset for MLLM evaluation containing 517 examples, each consisting of a text question paired with one or more images. 
We generate captions for each image using a captioning model (Molmo-7B)
and then extract embeddings for each image and its corresponding caption using an MLLM (also Molmo-7B) and use them to compute the \alignmentscore, which peaks in layer 22 (out of 28 layers; see \cref{fig:difference-cosine-sim-aligned-shuffled} green curve). 

To select layers aligned between text and audio, we use the audio version of the Alpacaeval dataset from VoiceBench,
a dataset of 636 audio-transcript pairs. 
We extract embeddings from each layer of an MLLM (LLaMA-Omni 8B) and use them to compute the \alignmentscore, which peaks at layer 21 (out of 32 layers; see \cref{fig:difference-cosine-sim-aligned-shuffled} purple curve).

\smallskip
Overall, we see that LLMs and MLLMs generate representations that are shared across languages and modalities.

\subsection{Training and evaluating the harmfulness classifier}

\subsubsection{Setup for multilingual text attacks}

\paragraph{\toolname classifier.} 
Following \cref{subsec:layer_selection},
we build a classifier that takes as input embeddings from layer 57 of Llama3.3-70B-Instruct.
As training data, we randomly select 2,800 examples from the Aegis AI Content Safety dataset, balancing the benign and harmful classes.
Notably, this dataset is about 18$\times$ smaller than the training data used by our baseline methods. 
We translate these English examples to 52 other natural languages (via the Google Translate API) and 20 cipher languages (using fixed rules), totaling 73 languages.
We train \toolname using only half the languages (see list in \cref{sec:languages-used}).

\paragraph{Baselines. }
We compare to many popular guard models (see \cref{tab:model_details}) middle row.
Notably, \textit{DuoGuard} and \textit{PolyGuard} were trained to detect harmful prompts across multiple languages. 
For a more direct comparison, we also compare to finetuned versions of \textit{DuoGuard} and \textit{PolyGuard} using the same 37 languages we use to train \toolname;
Following the original \textit{PolyGuard} paper, we finetuned these models using LoRA \citep{lora-paper} for all linear layers with rank 8 and alpha 16 for one epoch with a learning rate of $2e-4$.

\paragraph{Datasets.}
We evaluate on several common text attack benchmarks (see \cref{tab:dataset_details}).
We additionally evaluate on three benchmarks from CodeAttacks that transform a harmful query as a list, a stack, or as a string in a Python program, obfuscating the harmfulness. 
For evaluation in this setup, we transform the harmful prompts from AdvBench and benign prompts from Toxicity Jigsaw datasets in the three code formats and subsample the Toxicity Jigsaw dataset to be of the same size as Advbench.
Note that for this experiment, we only trained \toolname on the English subset of the training dataset.

\input{tables/models-used}

\subsubsection{Setup for vision attacks}
\label{sec:setup-images}

\paragraph{\toolname classifier.} 
Following \cref{subsec:layer_selection}, we build a classifier that takes as input embeddings from layer 22 of Molmo-7B.
As training data, we use 2000 image-query pairs randomly sampled from the JailBreakV-28K dataset and 1024 image-query pairs sampled from the VLSafe dataset as the harmful datapoints and 517 image-query from the MM-Vet v2 dataset as the benign datapoints. 

\paragraph{Baselines.}
We compare to guard models that take an image or image-text pair and output a binary harmfulness classification (see \cref{tab:model_details} bottom row).
We train \textit{VLMGuard} on the same training data as \toolname.

\paragraph{Datasets.}
We evaluate detecting image/text attacks using several datasets (see \cref{tab:dataset_details}).
FigStep and MML Safebench are typographic attacks that embed a harmful prompt in an image. MML Safebench further encrypts a harmful prompt in several variants, such as rotation, mirror images, word replacement, and with base64 encoding.
Hades and Safebench consist of images and text queries where the text itself is harmful. 
MM-safetybench, RTVLM, and VLSBench consist of an image and a query where the text query is seemingly benign, but when combined with the respective image, it is harmful (e.g. see \Cref{fig:main-figure}).

\subsubsection{Setup for audio attacks}
\label{sec:setup-audio}

\paragraph{\toolname classifier.} 
Following \cref{subsec:layer_selection},
we build a classifier that takes as input embeddings from layer 21 of Llama-Omni-8B.
We train the classifier on the English portion of the training data we use for the text setting, by using a text-to-speech model to convert the text into audio. We use the open-source Kokoro model as the text-to-speech model. 

\paragraph{Baselines.}
We are unaware of any existing models for detecting harmful audio input.
The most relevant approach, SpeechGuard \citep{peri-etal-2024-speechguard} adds noise as a defense against potentially harmful audio inputs but does not directly classify harmfulness.
To contextualize our results for audio benchmarks, we compare performance to guard models that directly classify the raw text present in the audio (\toolname and LlamaGuard3).

\paragraph{Datasets.}
We use the two audio benchmarks (see \cref{tab:dataset_details} bottom row).
We also evaluate on several text jailbreak benchmarks using Kokoro to convert them from text to speech: HB, FQ, Simple Safety Tests, SaladB, and TJS. 
We use Kokoro for generating text-to-speech versions.

%% file: tables/datasets-used.tex
\begin{table*}
    \small
    \centering
    \resizebox{0.8\textwidth}{!}{%
    \begin{tabular}{llllr}
        \toprule
         & Dataset name & Citation & HuggingFace ID & Number of examples\\
         \midrule
          \parbox[c]{2mm}{\multirow{3}{*}{\rotatebox[origin=c]{90}{General}}}
         & Flores200 & \cite{nllbteam2022-languageleftbehindscaling} & \texttt{Muennighoff/flores200} & 997\\
         & MM-Vet v2 & \cite{yu2024-mmvetv2-dataset} & \texttt{whyu/mm-vet-v2} & 517 \\
         & SST-2 & \citep{socher-etal-SSt-5-dataset} & \texttt{stanfordnlp/sst2} & 1000 \\ \midrule
          
          \parbox[c]{2mm}{\multirow{15}{*}{\rotatebox[origin=c]{90}{Text}}}
         & Aegis AI Content Safety Dataset & \cite{ghosh2024aegis2} & \texttt{nvidia/Aegis-AI-Content-Safety-Dataset-1.0} & 10,800 \\ 
         & MultiJail  & \cite{deng2024-multilingual-attack-firstpaper} & \texttt{DAMO-NLP-SG/MultiJail} & 315 \\
         & Xsafety  & \cite{wang2024-multilingualsafety} & \texttt{ToxicityPrompts/XSafety} & 28,000 \\
         & RTP-LX  & \cite{rtplx-dataset} & \texttt{ToxicityPrompts/RTP-LX} & 30,300 \\
         & AyaRedTeaming  & \cite{aya-redteaming-dataset} & \texttt{CohereLabs/aya_redteaming} & 2662 \\
         & Thai Toxicity tweets & \cite{thai-toxicity-tweets} & \texttt{tmu-nlp/thai_toxicity_tweet} & 3,300 \\
         & Ukr Toxicity & \cite{ukr-toxicity-dataset} & \texttt{ukr-detect/ukr-toxicity-dataset}  & 5,000 \\
         & HarmBench (HB) & \citep{mazeika2024-harmbench} & \texttt{walledai/HarmBench} & 400 \\
         & Forbidden Questions (FQ) & \citep{forbidden-questions-dataset} & \texttt{TrustAIRLab/forbidden_question_set} & 390 \\
         & Simple Safety Tests & \citep{simple-safety-tests} & \texttt{walledai/SimpleSafetyTests} & 100 \\
         & SaladBench (SaladB) & \citep{salad-bench-dataset} & \texttt{walledai/SaladBench} & 26,500 \\
         & Toxicity Jigsaw (TJS) & \cite{toxicity-jigsaw} & \texttt{Arsive/toxicity_classification_jigsaw} & 26,000 \\
         & Toxic Text & \citep{toxic-text-dataset} & \texttt{nicholasKluge/toxic-text} & 41,800 \\
         & AdvBench & \citep{advbench-dataset} & \texttt{walledai/AdvBench} & 520 \\ 
         & CodeAttack & \citep{ren-2024-codeattack-jailbreak} & \texttt{https://github.com/AI45Lab/CodeAttack} & 3120 \\
         \midrule
          
          \parbox[c]{2mm}{\multirow{9}{*}{\rotatebox[origin=c]{90}{Vision}}} 
         & JailBreakV-28K & \citep{luo2024jailbreakv28k} & \texttt{JailbreakV-28K/JailBreakV-28k} & 8,000 \\
         & VLSafe & \citep{chen2024-vlsafe-dataset} & \texttt{YangyiYY/LVLM_NLF} & 1,110 \\
         & FigStep & \citep{figstep-dataset-image-jailbreak} & \texttt{https://github.com/wangyu-ovo/MML} & 500 \\
         & MML SafeBench & \citep{mml-safebench-image-jailbreak} & \texttt{https://github.com/wangyu-ovo/MML} & 2,510 \\
         & Hades & \citep{hades-dataset-image-jailbreak} & \texttt{Monosail/HADES} & 750 \\
         & SafeBench & \citep{another-safebench-dataset-image-jailbreaks} & \texttt{Zonghao2025/safebench} & 2,300 \\
         & MM SafetyBench & \citep{mm-safetybench-image-jailbreak} & \texttt{PKU-Alignment/MM-SafetyBench} & 1680 \\
         & RedTeamVLM & \citep{li2024-RedTeamVLM-dataset} & \texttt{MMInstruction/RedTeamingVLM} & 200 \\
         & VLSBench & \citep{vlsbench-dataset-image-jailbreak} & \texttt{Foreshhh/vlsbench} & 2,240 \\ \midrule
          
          \parbox[c]{2mm}{\multirow{2}{*}{\rotatebox[origin=c]{90}{Audio}}} 
         & VoiceBench (Alpacaeval) & \cite{advbench-audio-jailbreaks} & \texttt{hlt-lab/voicebench}  & 636 \\
         & AIAH & \citep{achilles-dataset-audio-jailbreaks} & \texttt{https://github.com/YangHao97/RedteamAudioLMMs} & 350 \\
         \bottomrule
    \end{tabular}
    }
    \caption{Details of datasets used for training and evaluation.
    Some of the text datasets are inherently multilingual : MultiJail (10 languages), XSafety (10 languages), RTP-LX (28 languages), Aya RedTeaming (8 languages), Thai Toxicity tweets (prompts in Thai), and Ukr Toxicity (prompts in Ukrainian).
    The remaining text datasets are English-only, and were translated  to 72 other languages (52 natural and 20 cipher):
    HarmBench (HB), Forbidden Questions (FQ), Simple Safety Tests, SaladBench (SaladB), Toxicity Jigsaw (TJS), Toxic Text, and AdvBench.
    }
    \label{tab:dataset_details}
\end{table*}

%% file: tables/models-used.tex
\begin{table*}
    \small
    \centering
    \resizebox{0.8\textwidth}{!}{%
    \begin{tabular}{llllr}
        \toprule
         & Model name & Citation & HuggingFace ID & Rough Parameter Count\\
         \midrule
    \parbox[c]{2mm}{\multirow{4}{*}{\rotatebox[origin=c]{90}{General}}}
    & Llama3.3-70B-Instruct & \citep{grattafiori2024llama} & \texttt{meta-llama/Llama-3.3-70B-Instruct} & 70B\\
    & Molmo-7B &  \citep{deitke2024-molmo-model} & \texttt{allenai/Molmo-7B-D-0924} & 7B\\
    & LLaMA-Omni 8B & \citep{fang2025-llama-omni-model} & \texttt{ICTNLP/Llama-3.1-8B-Omni} & 8B \\
    & Kokoro & \citep{hexgrad_kokoro_82m_2025} & \texttt{hexgrad/Kokoro-82M} & 82M \\
         \midrule
    
    \parbox[c]{2mm}{\multirow{10}{*}{\rotatebox[origin=c]{90}{Text}}}
    & LlamaGuard 1 & \cite{inan2023-llama-guardrails} & \texttt{meta-llama/LlamaGuard-7b} & 7B  \\
    & LlamaGuard 2 & \cite{inan2023-llama-guardrails} & \texttt{meta-llama/Meta-Llama-Guard-2-8B} & 8B  \\
    & LlamaGuard 3 & \cite{inan2023-llama-guardrails} & \texttt{meta-llama/Llama-Guard-3-8B} & 8B \\
    & AegisGuard Permissive & \cite{ghosh2024-aegis-safety-guard} & \texttt{nvidia/Aegis-AI-Content-Safety-LlamaGuard-Permissive-1.0} & 7B  \\
    & AegisGuard Defensive & \cite{ghosh2024-aegis-safety-guard} & \texttt{nvidia/Aegis-AI-Content-Safety-LlamaGuard-Defensive-1.0} & 7B  \\
    & WildGuard & \cite{Han2024-wildguardmix-dataset} & \texttt{allenai/wildguard} & 7B \\
    & HarmBench (mistral) & \cite{mazeika2024-harmbench} & \texttt{cais/HarmBench-Mistral-7b-val-cls} & 7B \\
    & HarmBench (llama) & \cite{mazeika2024-harmbench} & \texttt{cais/HarmBench-Llama-2-13b-cls} & 13B \\
    & DuoGuard & \cite{duo-guard-multilingual} & \texttt{DuoGuard/DuoGuard-1B-Llama-3.2-transfer} & 1B \\
    & PolyGuard & \cite{polyguard-model} & \texttt{ToxicityPrompts/PolyGuard-Qwen} &  7B \\ \midrule
    
    \parbox[c]{2mm}{\multirow{3}{*}{\rotatebox[origin=c]{90}{Vision}}}
    & Llama Guard 3 Vision & \citep{chi2024-llamaguard3-vision} & \texttt{meta-llama/Llama-Guard-3-11B-Vision} & 11B \\
    & VLMGuard & \citep{du2024-vlmguard} & \texttt{---} & 2.2M \\
    & LLavaGuard & \citep{helff2025-llavaguard}  & \texttt{AIML-TUDA/LlavaGuard-7B-hf} & 7B \\ 
         \bottomrule
    \end{tabular}
    }
    \caption{Model and baseline details.}
    \label{tab:model_details}
\end{table*}

%% file: 04_results.tex
\section{Results}

\begin{table*}
    \centering
    \small
    \makebox[\textwidth]{
    \resizebox{1.05\textwidth}{!}{%
    \input{tables/multilingual-set1}
    }
}
    \makebox[\textwidth]{
    \resizebox{1.03\textwidth}{!}{%
\input{tables/multilingual-set2}
    }}
    \makebox[\textwidth]{
    \resizebox{1.03\textwidth}{!}{%
    \input{tables/multilingual-set3}
    }
}
\caption{Accuracy of detecting harmful prompts for text attack benchmarks that are (A) multilingual benchmarks, (B) English translated to 73 languages, and (C) English translated to languages not seen at training time.
In all settings, \toolname achieves the highest performance.
\Cref{tab:high-low-cipher-performance} further stratifies these results by high-resource, low-resource, and cipher languages.
}
\label{tab:results-benchmark-set-1}
\end{table*}

\begin{table*}
    \centering
    \small
    \makebox[\textwidth]{
    \resizebox{1\textwidth}{!}{%
\input{tables/vision-set1}
}
}
    \label{tab:results-image-jailbreaks-2}
    \makebox[\textwidth]{
    \resizebox{1\textwidth}{!}{%
\input{tables/vision-set2}}
}
\caption{Accuracy of detecting harmful queries in multimodal benchmarks for (A) image-query pairs and (B) typographed images with encrypted text.
\toolname achieves the highest performance for both kinds of benchmarks.}
\label{tab:results-image-jailbreaks-1}
\end{table*}

\begin{table*}
    \small
    \centering
    \makebox[\textwidth]{
    \resizebox{1.05\textwidth}{!}{%
    \input{tables/audio}
    }}
    \caption{Accuracy of detecting harmful queries in audio.
    \toolname is able to detect harmful audio inputs with high accuracy across all benchmarks. Since there are no baselines for detecting harmful prompts in audio, we compare the performance against \toolname's and LlamaGuard3 when the same benchmarks are provided as text in English. }
    \label{tab:results-audio-jailbreaks}
\end{table*}

\paragraph{Defending against multilingual text attacks}
\label{sec:results-multilingual}

\Cref{tab:results-benchmark-set-1} compares the accuracy of detecting harmful prompts for text benchmarks.
\Cref{tab:results-benchmark-set-1}(A) shows results for multilingual benchmarks, where \toolname achieves the highest accuracy (86.36\%) compared to the baselines, and achieves new state-of-the-art performance for 3 benchmarks: MultiJail, RTP-LX, and AyaRedTeaming.
The strongest baseline is Polyguard, which yields an average accuracy of 83.19\%, despite being trained on a much larger dataset (1.91M examples for Polyguard versus 103K examples for \toolname). 
In benchmarks that were translated from English to various other languages, including cipher languages, we again see that \toolname achieves the highest accuracy (\Cref{tab:results-benchmark-set-1}(B)).
Finally,
\Cref{tab:results-benchmark-set-1}(C) shows that \toolname outperforms finetuned versions of DuoGuard and Polyguard on unseen languages, demonstrating that \toolname can outperform methods that were trained specifically for multilingual harmfulness classification.

\paragraph{Defending against image-based attacks}
\label{sec:results-image}

\Cref{tab:results-image-jailbreaks-1} 
shows the accuracy of detecting harmful image and text prompts
for
(A) pairs consisting of images and text queries, where either the image or both the image and query can be harmful
and (B) typographic images with various encryptions.
\toolname achieves the highest performance for both sets of benchmarks (\textbf{95.44\%} and \textbf{79.76\%}) while being trained using only about 3500 image-query pairs (compared to about 5500 datapoints used by LlavaGuard).
The only benchmark where \toolname fails to detect harmful prompts is MML Base64, which consists of typographed images of prompts encrypted using base64 encoding.

\paragraph{Defending against audio-based attacks}
\label{sec:results-audio}

\Cref{tab:results-audio-jailbreaks} shows the accuracy of detecting harmful audio prompts. \toolname detects harmful audio input with high accuracy across all benchmarks. 
As we are not aware of any existing defenses for audio jailbreaks,
we compare against 
\toolname and LlamaGuard3's accuracy in detecting harmful prompts when the same inputs are provided in English text.
The accuracy \toolname achieves in detecting harmful audio inputs is similar to or higher than its performance for detecting harmful text inputs. 

\begin{figure}[h!]
    \centering
    \includegraphics[width=0.9\linewidth]{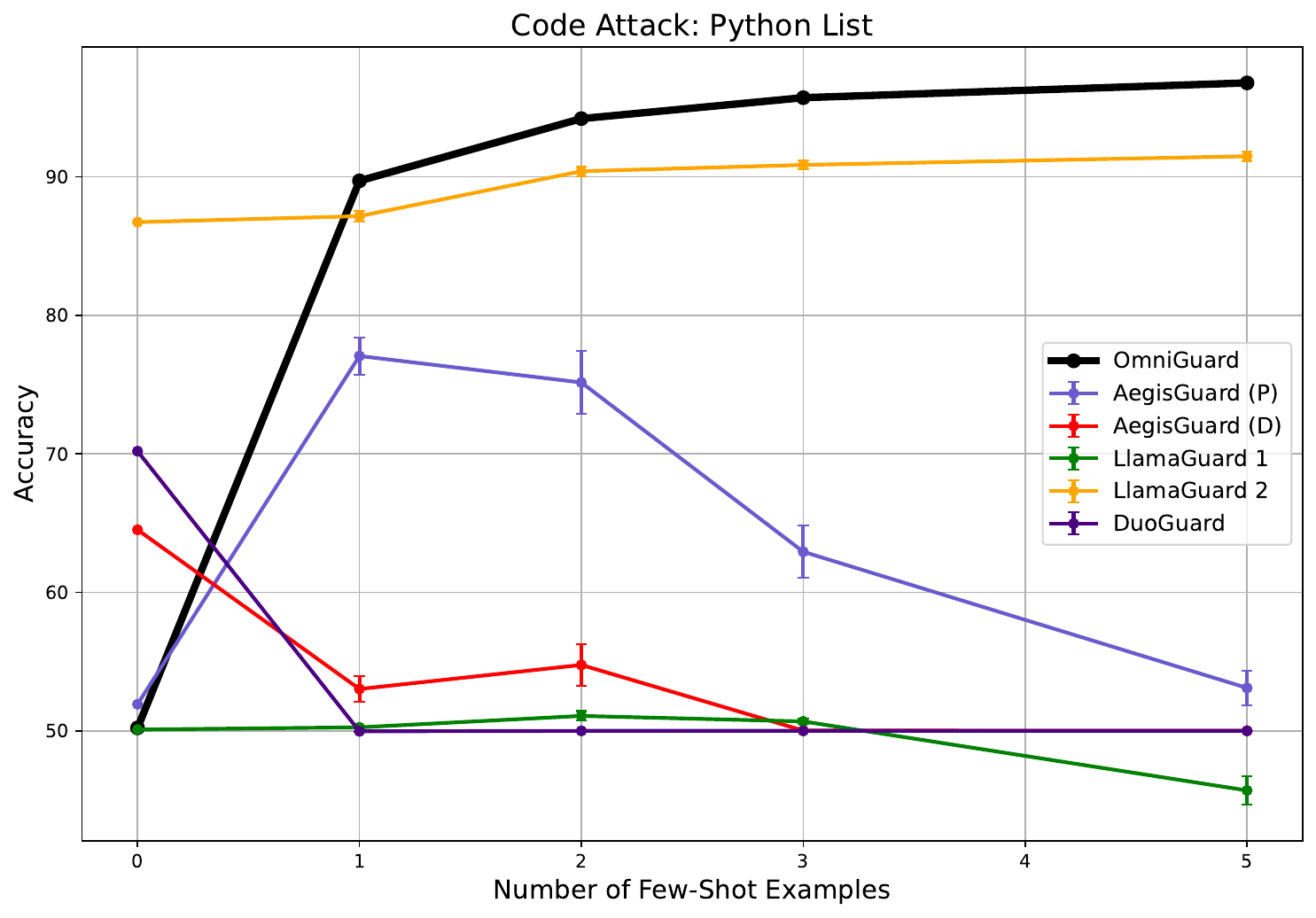}
    \vspace{1mm}
    \includegraphics[width=0.9\linewidth]{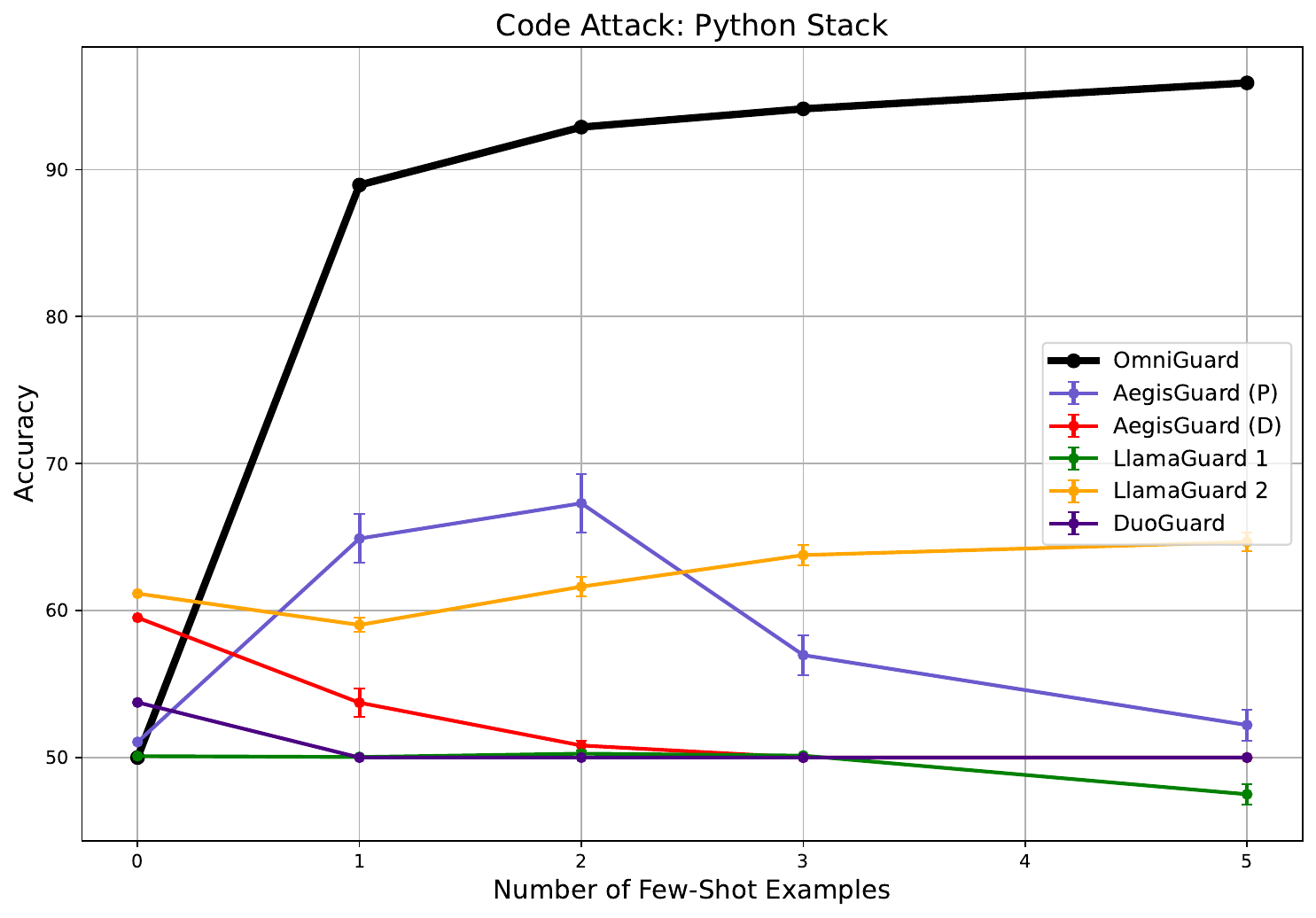}
    \vspace{1mm}
    \includegraphics[width=0.9\linewidth]{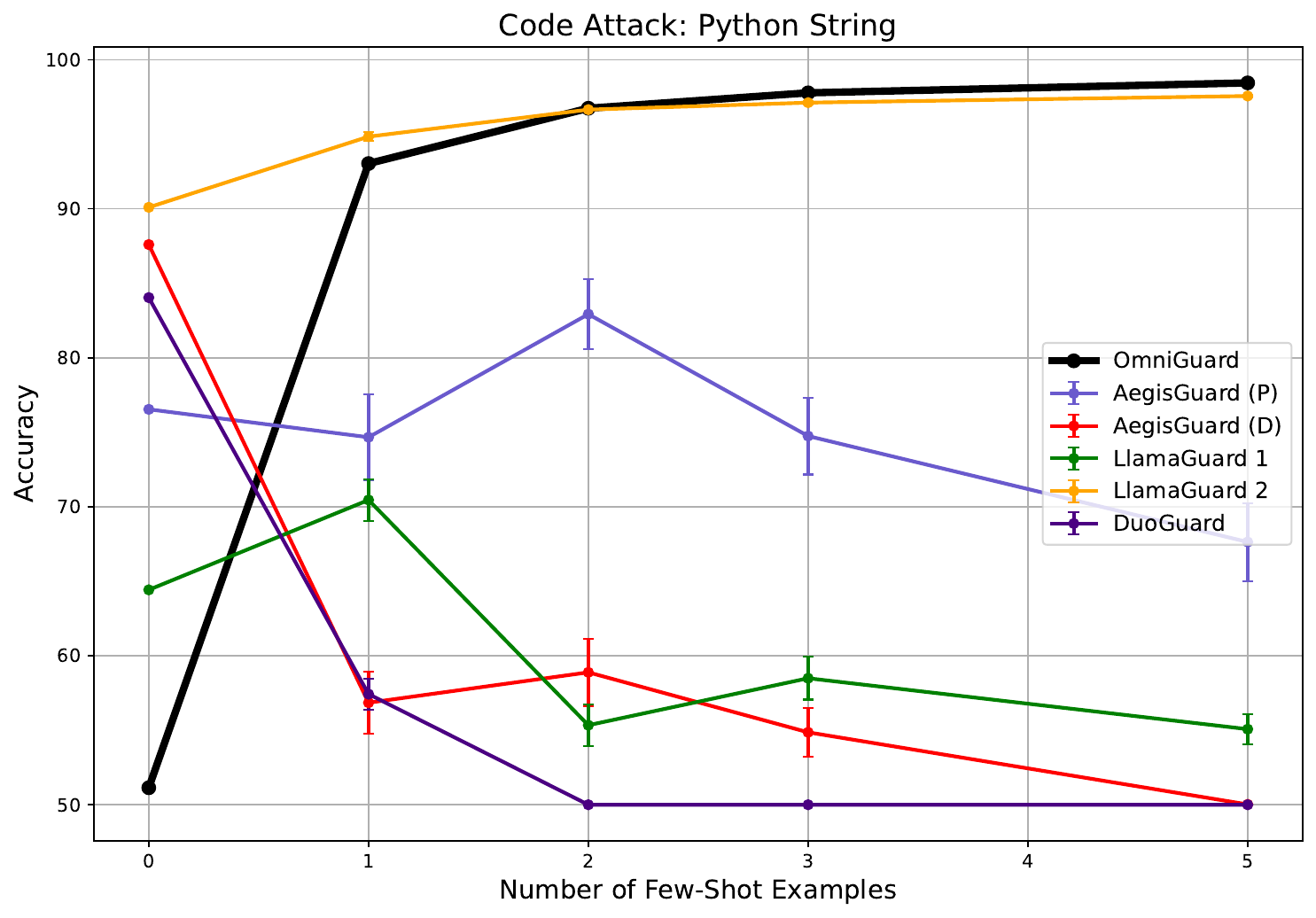}
    \caption{Accuracy of detecting harmful prompts in a few-shot setting.
    As few-shot examples are provided, \toolname quickly achieves near-perfect accuracy, despite the attacks being quite different from its training data (e.g. without any few-shot examples, \toolname's accuracy is close to 50\%
    ).
    In contrast, the guard model baselines improve their accuracy slowly in a few-shot setting, despite sometimes having seen similar code attacks in their training data.
    Accuracies are averaged over 50 random sets of few-shot examples;
    error bars show the standard error of the mean. 
    }
    \label{fig:few-shot-experiments}
\end{figure} 

\paragraph{Data-efficient adaptation}
We also evaluate the accuracy of \toolname and baselines in adapting to out-of-distribution code attacks given very few samples.
In this setting, some prior work has speculated that guard models may be very data efficient, as they can make use of few-shot examples in-context~\citep{inan2023-llama-guardrails}.
However, we find that baseline guard models generally struggle to rapidly adapt to this setting given few-shot examples (\Cref{fig:few-shot-experiments}).\footnote{Note that we omit baseline guard models that achieve 90\% accuracy or greater without any few-shot examples, as their training data likely explicitly includes code attacks.}
In contrast, \toolname is able to rapidly achieve close to 100\% accuracy for all three benchmarks by updating its lightweight parameters using less than five examples.

%% file: tables/multilingual-set1.tex
\begin{tabular}{ll|cccccc|c}
\toprule
& & MultiJail & Xsafety & RTP-LX & Aya RedTeaming & Thai Tox & Ukr Tox & Avg. \\ \midrule
 \parbox[c]{2mm}{\multirow{12}{*}{\rotatebox[origin=c]{90}{(A)}}}
 \parbox[c]{2mm}{\multirow{12}{*}{\rotatebox[origin=c]{90}{Multilingual text benchmarks}}} & LlamaGuard 1 & 39.27 & 57.01 & 48.66 & 54.49  & 41.31 & 53.99 & 49.12\\
 & LlamaGuard 2 & 48.69 & 52.66 & 34.69 & 58.58 & 42.86 & 51.79 & 48.21 \\
 & LlamaGuard 3 & 66.87 & 64.34 & 45.57 & 63.83 & 46.73 & 51.79 & 56.52 \\
 & AegisGuard (P) & 61.49 & 79.78 & 75.07 & 78.88 & 56.09 & 65.75 & 69.51 \\
 & AegisGuard (D) & 79.71 & 90.77 & 92.17 & 89.78 & 63.34 & 67.95 & 80.62 \\
 & WildGuard & 42.55 & 71.23 & 71.94 & 61.45 & 40.42 & 55.03 & 57.10 \\
 & HarmBench (llama) & 0.22 & 0.14 & 0.0 & 0.03 & 39.04 & 50.1 & 14.92 \\
 & HarmBench (mistral) & 2.4 & 5.65 & 5.14 & 7.39 & 40.42 & 50.55 & 18.59 \\
 & MD-Judge  & 25.78 & 53.58 & 66.46 & 46.20 & 39.48 & 53.89 & 47.56 \\
 & DuoGuard & 39.20 & 63.42 & 66.57 & 61.80 & 45.63 & 50.75 & 54.56 \\
 & PolyGuard & 82.00 & \textbf{96.41} & 83.86 & 90.34 & \textbf{70.43} & \textbf{76.07} & 83.19 \\
 & \toolname & \textbf{93.83} & 93.64 & \textbf{94.55} & \textbf{94.31} & 68.7 & 73.1 & \textbf{86.36} \\ \bottomrule
\end{tabular}

%% file: tables/multilingual-set2.tex
\begin{tabular}{ll|ccccccc|c}
\toprule
&  & HarmBench & FQ & SimpleST & SaladB & TJS & ToxText & AdvBench & Avg. \\ \midrule
\parbox[c]{2mm}{\multirow{12}{*}{\rotatebox[origin=c]{90}{(B)}}}
\parbox[c]{2mm}{\multirow{12}{*}{\rotatebox[origin=c]{90}{Translated text benchmarks}}} & LlamaGuard 1  & 32.47 & 23.75 & 34.32 & 23.27 & 62.49 & 65.55 & 34.39 & 39.46 \\
 & LlamaGuard 2  &  57.19 & 43.72 & 50.71 & 34.54 & 58.21 & 62.17 & 56.95 & 51.93  \\
 & LlamaGuard 3 & 70.02 & 53.25 & 67.81 & 46.30 & 62.33 & 70.87 & 70.26 & 62.98 \\
 & AegisGuard (P)  &  62.16 & 43.01 & 56.55 & 44.92 & 73.69 & 72.80 & 62.12 & 59.32  \\
 & AegisGuard (D)  & 88.53 & 76.67 & 87.64 & 78.27 & 71.38 & 68.72 & \textbf{90.77} & 80.28  \\
 & WildGuard  &  33.64 & 31.20 & 33.90 & 27.37 & 66.61 & 67.27 & 39.98 & 42.85  \\
 & HarmBench (llama)  & 0.03 & 0.11 & 0.01 & 0.07 & 48.62 & 49.97 & 0.01 & 14.12 \\
 & HarmBench (mistral)  & 2.32 & 1.75 & 2.04 & 1.66 & 50.53 & 50.69 & 1.7 & 15.81  \\
 & MD-Judge   &  16.19 & 12.11 & 22.29 & 13.81 & 65.34 & 64.26 & 25.67 & 31.38  \\
 & DuoGuard  &  20.44 & 44.36 & 28.79 & 36.88 & 68.57 & 69.07 & 28.58 & 42.38  \\
 & PolyGuard  & 66.22 & 56.05 & 62.53 & 54.88 & \textbf{78.34} & \textbf{76.52} & 67.96 & 66.07 \\
 & \toolname & \textbf{89.13} & \textbf{89.57} & \textbf{89.62} & \textbf{87.30} & 76.68 & 75.07 & 86.59 & \textbf{84.85} \\ \bottomrule
\end{tabular}%

%% file: tables/multilingual-set3.tex
\begin{tabular}{ll|ccccccc|c}
\toprule
& & HarmBench & FQ & SimpleST & SaladB & TJS & ToxText & AdvBench & Avg. \\ \midrule
 \parbox[c]{2mm}{\multirow{3}{*}{\rotatebox[origin=c]{90}{(C)}}}
 \parbox[c]{2mm}{\multirow{3}{*}{\rotatebox[origin=c]{90}{Unseen}}} \parbox[c]{1mm}{\multirow{3}{*}{\rotatebox[origin=c]{90}{langs.}}}& FT DuoGuard &  23.59 & 39.08 & 28.14 & 33.29 & 54.1 & 53.23 & 28.29 & 37.1  \\
 & FT PolyGuard  & 72.45 & 79.84 & 76.81 & 76.85 & \textbf{74.07} & \textbf{72.33} & 73.55 & 75.13 \\
 & \toolname & \textbf{86.51} & \textbf{86.65} & \textbf{86.42} & \textbf{85.01} & 72.82 & 71.44 & \textbf{84.29} & \textbf{81.88} \\ \bottomrule
\end{tabular}%

%% file: tables/vision-set1.tex
    \begin{tabular}{ll|cccccc|c}
      \toprule
      & & Hades  & VLSBench & MM-SafetyBench & SafeBench & RTVLM & FigStep  & Avg. \\ \midrule
      \parbox[c]{2mm}{\multirow{4}{*}{\rotatebox[origin=c]{90}{(A)}}}
      \parbox[c]{2mm}{\multirow{4}{*}{\rotatebox[origin=c]{90}{Image}}} \parbox[c]{2mm}{\multirow{4}{*}{\rotatebox[origin=c]{90}{+Query}}} & Llama3 Vision GRD     &  76.00 & 3.97     & 31.90          & 68.40     & 56.50 & 47.40 & 47.36  \\
      & VLMGuard              &  98.00 & 74.56    & 92.20          & 73.90     & \textbf{94.00} & 99.80  & 88.74 \\
      & LLavaGuard            &  23.73 & 42.08    & 10.95          & 12.10     & 18.50 & 3.40  & 18.46 \\
      & \toolname             & \textbf{100.00} & \textbf{92.24}    & \textbf{99.82}          & \textbf{91.60}     & 89.00 & \textbf{100.00} & \textbf{95.44} \\
      \bottomrule
    \end{tabular}

%% file: tables/vision-set2.tex
    \begin{tabular}{ll|ccccc|c}
      \toprule
      &          & MML Rotate & MML Mirror & MML W.R. & MML Q.R.  & MML Base64     & Avg.  \\ \midrule
      \parbox[c]{2mm}{\multirow{4}{*}{\rotatebox[origin=c]{90}{(B)}}} 
      \parbox[c]{2mm}{\multirow{4}{*}{\rotatebox[origin=c]{90}{\scriptsize Typographed}}} \parbox[c]{2mm}{\multirow{4}{*}{\rotatebox[origin=c]{90}{\scriptsize image}}} & Llama3 Vision GRD     & 83.20      & 68.00      & 96.40    & 25.40     & \textbf{98.80} & 74.36    \\
      & VLMGuard              & 6.80       & 21.00      & \textbf{100.0}    & 86.20     & 0.20           & 42.84    \\
      & LLavaGuard            & 0.00       & 0.00       & 0.00     & 11.40     & 0.00           & 2.28    \\
      & \toolname             & \textbf{100.0}      & \textbf{100.0}      & 99.60    & \textbf{98.80}  & 0.40 & \textbf{79.76} \\
      \bottomrule
    \end{tabular}

%% file: tables/audio.tex
\begin{tabular}{l|ccccccccc}
  \toprule
  & AIAH & SafeBench (M) & SafeBench (F) & HB & FQ & SimpleST  &  SaladB   & TJS & AdvBench \\
  \midrule
  \toolname (Audio) & \textbf{91.14} & \textbf{94.4} & \textbf{93.8} & \text{95.98} & \text{90.42} & \text{97.0}  &    \textbf{94.21}     & \text{82.03}  & \textbf{98.85} \\ \midrule
  \toolname (text-en) & - & - & - & 92.0  & \textbf{93.3}  & 93.0  & 90.2  & \textbf{93.2}  & 90.0  \\
  LlamaGuard3 (text-en) & - & - & - & \textbf{97.32}  & 78.75  & \textbf{99.0}  & 67.03  & 72.16  & 98.07 \\ 
  \bottomrule
\end{tabular}

%% file: 05_analysis-multilingual.tex
\section{Analysis}
\label{sec:analysis}

\paragraph{Effect of \alignmentscore-based layer selection.}
We perform ablation experiments to determine the effect of selecting the appropriate layer for training the \toolname classifier.
For the text-only model, we compare the \alignmentscore-selected layer (57) to 3 other layers (layer 10, layer 75, and the last layer) when used for a set of toxicity prediction tasks.
\Cref{tab:results-layer-ablation} shows that the representations from the layer with the highest \textit{\alignmentscore} result in significantly better harmfulness classification accuracy, improving between 5\% and 14\% compared to the other layers. We show ablation over more layers in \Cref{tab:results-layer-ablation-merged}. 

\input{tables/mutilingual-layer-ablation}

\paragraph{\toolname's efficiency}
\toolname is highly efficient at inference time because it re-uses the internal representations of the main LLM that is already processing the user query for generation.
Therefore, its compute time is only that of a lightweight multilayer perceptron, making it much faster than baseline guard models (note that this does limit \toolname to only work when the generation model is open-source, so embeddings can be extracted).
\Cref{tab:inference-time-comparison} shows the inference time required by various guard models to predict the harmfulness of prompts in the AdvBench dataset in English, translated to Spanish, French, Telugu, and base64 encoding. \toolname is the fastest and is about \textbf{120$\times$ faster} than the fastest baseline (DuoGuard).
Inference time as measured on a machine with 1 L40 GPU, 4 CPUs, and 50 GB RAM. 

\begin{table}
    \centering
    \small
    \begin{tabular}{lr}
    \toprule
    Guard Method   & Inference Time (s) $\downarrow$ \\ \midrule
    LlamaGuard 3   &  87.25     \\
    AegisGuard (D) &  152.26    \\
    WildGuard      &  306.14      \\
    MD-Judge       &  128.26     \\
    DuoGuard       &  4.85     \\
    PolyGuard      &  409.90    \\
    \toolname      &  \textbf{0.04}    \\  
    \bottomrule
    \end{tabular}
    \caption{Average inference time required for harmfulness prediction on the AdvBench dataset (averaged over 5 languages). \toolname is about 120$\times$ faster than the fastest baseline (DuoGuard).}
     \label{tab:inference-time-comparison}
\end{table}

\paragraph{Performance comparison across base LLMs}
We compare \toolname's accuracy when using different base LLMs in \Cref{tab:results-llm-comparison}. We trained the classifiers on the layers with the best U-scores for each model. We find that the average accuracy for the moderator model trained using smaller LLMs is lower than the moderator model trained using the larger Llama3.3-70B-Instruct model. 

\begin{figure}
    \centering
    \includegraphics[width=\columnwidth]{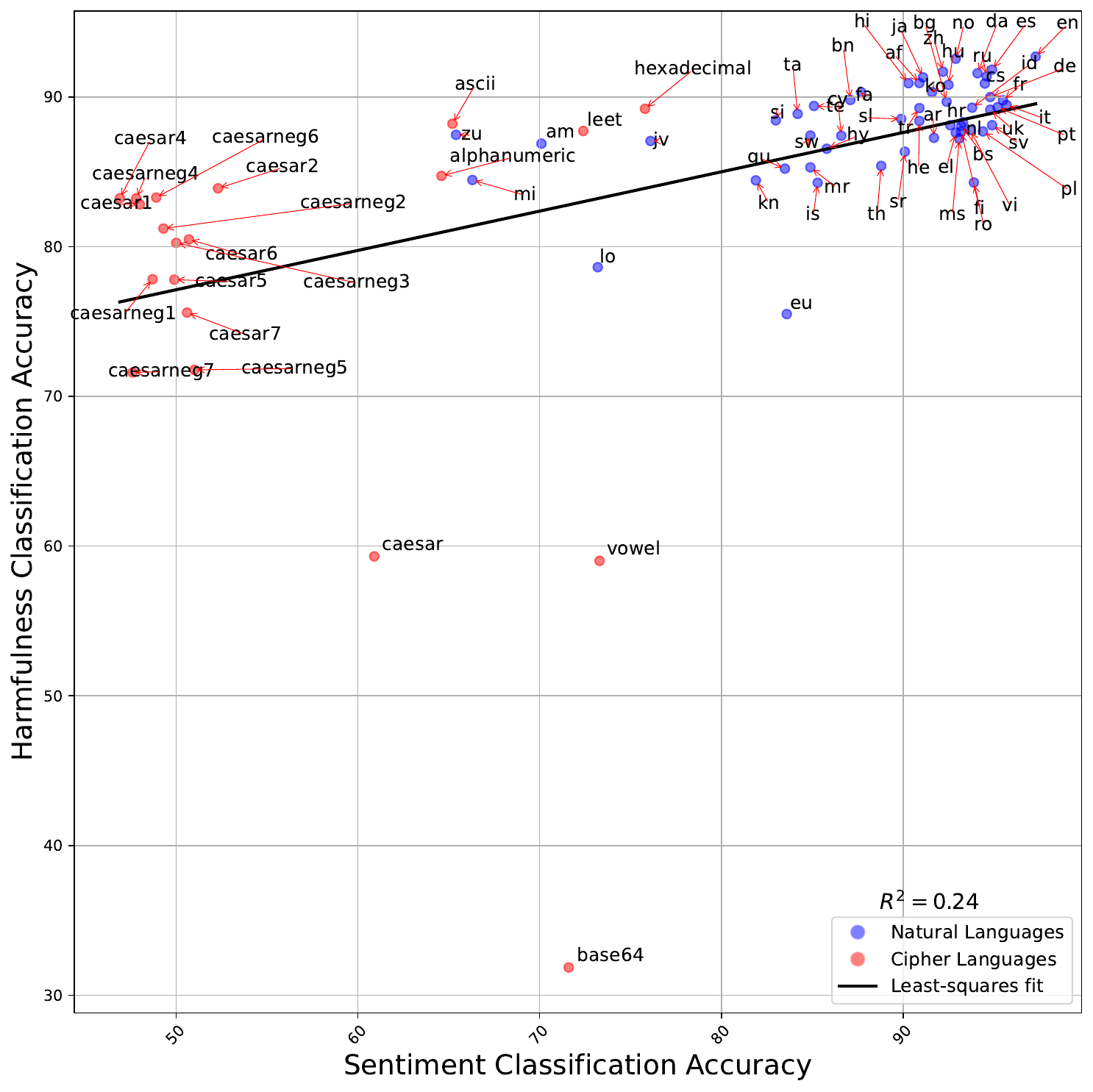}
    \caption{Comparison of accuracy of classifying sentiments in various languages compared to detecting harmful prompts in those languages using \toolname. In both cases the LLM is Llama3.3-70B-Instruct. 
    }
    \label{fig:scatter-plot-language-performance}
\end{figure}

\paragraph{Performance comparison across languages. }
We now analyze the harmfulness classification accuracy of \toolname by language, and compare it to the underlying LLM's sentiment classification accuracy for the same language (\cref{fig:scatter-plot-language-performance}).
We measure harmfulness classification accuracy using \toolname on all the datasets in \Cref{tab:results-benchmark-set-1} and sentiment classification accuracy using Llama3.3-70B-Instruct with zero-shot prompting on 72 translated versions of the SST-2 dataset (translated to all the languages we consider). 

We observe that the accuracies are generally correlated,
indicating that \toolname is able to defend well in languages for which the LLM is more coherent/susceptible to attack.
Unsurprisingly, the accuracies for natural languages are higher than the accuracies for cipher languages.
Nevertheless, harmfulness classification accuracy can be fairly high, even when sentiment classification accuracy is near chance (50\%).

%% file: tables/mutilingual-layer-ablation.tex
\begin{table}
    \centering
    \resizebox{\columnwidth}{!}{%
    \begin{tabular}{lcccc|c}
    \toprule
             & Thai Tox & Ukr Tox & TJS & ToxText & Avg. \\ \midrule
    Layer 10   &  62.1 &  65.5 &    66.95 &  61.89  &  64.42  \\ 
    Layer 75   & 65.2 &   66.4 &    70.72 & 65.79  & 68.26  \\
    Last Layer &  63.1 &  51.2 &    61.33 &  56.76  &  59.05  \\ 
    \alignmentscore selected layer (57)   & \textbf{68.7 } & \textbf{73.1}  & \textbf{76.8}     & \textbf{75.07}   &  \textbf{73.4} \\ 
    \bottomrule
    \end{tabular}%
    }
    \caption{\toolname's accuracy of detecting harmful prompts when trained using representations from different model layers. }
    \label{tab:results-layer-ablation}
\end{table}

%% file: mini_related_work.tex
\section{Related Work}

\paragraph{Jailbreak Attacks in LLMs}

Several techniques have recently emerged to attack or jailbreak LLMs.
Early techniques relied on manual effort and were very time-intensive \cite{shen2024-do-now-evaluating,andriushchenko2024-adaptive-attacks-best-asr-paper}.
Later techniques automated this process, e.g., \citet{zou2023-gcg-attacks,jones-another-gcg-attack,zhu2023-autodan-interpretable-gradient} proposed gradient-based approaches to identify inputs to jailbreak LLMs with white-box access. 
Another set of techniques start from a set of human written prompts and modify them using approaches like
genetic algorithms \citep{liu2024-autodan,lapid2024-opensesame-another-genetic,li2024-semanticmirror-another-genetic},
fuzzing \citep{yu2024-gpt-fuzzer-jailbreak},
or reinforcement learning \citep{chen2024-rl-jailbreak}
to automatically produce prompts for jailbreaking. 
Another set of techniques, use a helper LLM to generate prompts that attack a target LLM \citep{chao2024-PAIR-method,ding-2024-nested-attacks,mehrotra2024-tree-attacks-jailbreak}.
Finally,
\citet{wei2024-icl-jailbreak1,wang2023-icl-jailbreak2,anthropic-many-shot-jailbreak,pernisi2024-many-shot-jailbreak-italian} use simple in-context demonstrations to jailbreak the models by overcoming its safety training and \citet{russinovich2024-multi-turn-crescendo-attack,scale-ai-multi-turn-dataset} propose using multi-turn dialogues to jailbreak models.  

\paragraph{Multilingual Jailbreak Attacks}
Most of the aforementioned jailbreak techniques focus on attacks in English, against which significant defense exists both at the model and system level. 
To tackle this, a novel set of techniques have emerged that attack models using inputs in various languages or obfuscations that are able to bypass the safety guardrails. \citet{deng2024-multilingual-attack-firstpaper,yong2024-low-resource-languages-jailbreak-gpt4,wang2024-multilingualsafety,dan-roth-paper-guard-llms-on-multilingual-data,yoo2024-switch-between-languages,upadhayay2024-sandwich-attack-multilanguagemixture,song2024-multilingual-attacks} demonstrated that attacking models using mid and low resources languages led to higher attack success rates, compared to the case of attacking the model in high-resource languages like English. 

Going beyond natural languages, a newer set of works propose using cipher characters or languages to evade the safety filters, e.g., \citet{jin2024-JAMBench} propose interspersing cipher characters in between text, \citet{jiang2024-ascii-jailbreak} propose replacing the unsafe words with their ASCII art versions, and \citet{yuan2024-gpt-cipher-attack} propose prompting models in cipher languages like Morse, Atbash, Caesar.

\paragraph{Multimodal Jailbreak Attacks}
Using modalities apart from text aims to explore a completely new attack surface, like images or audios. Several recent works have shown that MLLMs remain vulnerable to being jailbroken when prompted with images or audios that have a harmful query (the same harmful query in text would be easily detected as harmful). \citet{mm-safetybench-image-jailbreak} show that using a prompt with a correlated image, e.g., using an image of a bomb when asking the model to answer the question: \textit{How to make a bomb?} is more likely to jailbreak a model than when using an uncorrelated image. \citet{vlsbench-dataset-image-jailbreak} argue that providing a harmful image with a benign query (see \Cref{fig:main-figure}) further increases the potential of jailbreaking the model. \citet{figstep-dataset-image-jailbreak} and \citet{mml-safebench-image-jailbreak} demonstrate jailbreaking models by simply typographically embedding harmful queries in an image. 

\paragraph{Safety moderation in LLMs}
Safety moderation in LLMs broadly fits into two categories: intrinsic and extrinsic. 
Intrinsic mechanisms include finetuning or RLHF training on an LLM \citep{bianchi2024-finetuning-for-safety,chen2024-finetuning-for-safety,yuan2024-refuse-feel-unsafe,rlhf-paper1,rlhf-paper2,dai2023-safe-rlhf}. 
Extrinsic safety mechanisms utilize external models to detect harmful inputs and responses; 
these models can either be simple filters or use guard models. \citet{jain2023-perplexity-defense1,alon2023-perplexity-defense2,hu2024-perplexity-defense3} propose using perplexity filtering for detecting harmful prompts. 
Guard models defend LLMs by training separate LLMs to detect harmful text (see \cref{tab:model_details}). 
Separately, a line of work has introduced interpretability methods to transparently expose safety concerns in LLMs~\citep{bereska2024mechanistic,singh2023augmenting,arditi2024refusal,benara2024crafting}.

A few other works also use internal model representations to defend against harmful inputs. However, they defer from our work: \toolname is a standalone content safety classification model while the other approaches like Jailbreak Antidote \citep{shen2025jailbreak-antidote} and AdaSteer \citep{zhao2025-ada-steer} directly change the internal representations to defend against harmful inputs. Therefore, \toolname is similar to other safety classification models like LlamaGuard and WildGuard. Additionally, \toolname can detect harmful prompts in multiple languages (both natural and cipher) and multiple modalities using the same method while the other approaches only defend against attacks in English text. \toolname is also extremely efficient in producing safety predictions while the other approaches take as much time as the inference of the underlying LLM, which can typically take several seconds to minutes.

\paragraph{Safety against multilingual attacks}

\textit{DuoGuard}~\citep{duo-guard-multilingual} and \textit{PolyGuard}~\citep{polyguard-model} are the two previous guard models that were specifically trained to defend against multilingual attacks.
DuoGuard uses a two-player RL-driven mechanism to generate harmful data in multiple languages and uses that to finetune a Llama3.2-1B model.
PolyGuard collects an extensive dataset of 1.91M samples of harmful and benign datapoints in 17 languages and uses that to finetune a Qwen-2.5-7B-Instruct model.

\paragraph{Safety moderation in MLLMs}
Relatively few works have tackled detecting harmful prompts in multimodal settings (see \cref{tab:dataset_details} and \cref{tab:model_details}).
\citet{chi2024-llamaguard3-vision} propose LlamaGuard3-11B-Vision (a finetuned version of Llama-3-11B-Vision) for detecting unsafe inputs in images and texts. \citet{du2024-vlmguard} and \citet{helff2025-llavaguard} propose other approaches for the same task. \toolname achieves higher accuracy in detecting harmful images and prompts compared to these approaches, and to the best of our knowledge is the first guard model for harmful audio inputs.

%% file: conclusions.tex
\section{Conclusions}
\label{sec:conclusions}

We propose \toolname, an approach for training a safety moderation classifier using the internal representations of an LLM or MLLM that are universally similar across languages and modalities. 
Our approach consists of two steps: first, we identify these universally similar representations and then we use them to train a harmfulness classifier. 
We find that \toolname accurately detects harmful prompts across languages, including low-resource languages as well as cipher languages, and also across modalities -- images and audios. 
We show that \toolname allows to train more efficient safety moderation classifiers (both in training time and in inference time) compared to standard guard models, and conclude that our approach is superior in both accuracy and efficiency across languages and modalities. 

%% file: limitations.tex
\section*{Limitations}
\label{sec:limitations}

While \toolname achieves state-of-the-art performance for detecting harmful prompts across languages and modalities, its performance depends on the underlying model. If the underlying model does not understand the language or an image or audio input, \toolname might not be able to detect if the input is harmful. However, this limitation is not unique to \toolname, and existing approaches suffer from the same limitation. 

Our approach also relies on the existence of universally similar representations, which we empirically found to exist across models and modalities. However, we did not exhaustively check all models and this assumption might not hold for models that we have not used in this work. 
Moreover, \toolname requires access to internal representations of a model, making it inapplicable to closed-source models.

Lastly, the results we report are based on a fixed set of evaluation datasets that are standard benchmarks used in the research area of AI safety moderation. While \toolname performs well across the datasets we experiment with, its performance in real-world settings might differ.

\textbf{Ethics.}
While this work seeks to mitigate the risks of LLM deployment in high-risk scenarios,
\toolname is not a perfect classifier and unexpected failures may allow for the harmful misuse of LLMs.

%% file: appendix.tex
\input{appendix-folder/languages-used}

\section{Datasets and models}
\setsectionlabel{B}
\label{sec:datasets_and_models}

\begin{table}
    \small
    \centering
    \makebox[\columnwidth]{
    \resizebox{1.05\columnwidth}{!}{%
    \input{tables/accuracy_by_resourceness}

    }}
    \caption{Accuracy of detecting harmful prompts stratified by high-resource natural, low-resource natural, and cipher languages. }
    \label{tab:high-low-cipher-performance}
\end{table}

\input{appendix-folder/best-layer-experiment-details}

\input{appendix-folder/toolname-architecture}

%% file: appendix-folder/languages-used.tex
\section{Languages Used in Our Approach}
\label{sec:languages-used}

We use the following languages in our experiments:
\begin{enumerate}
    \item Natural Languages: English, French, German, Spanish, Persian, Arabic, Croatian, Japanese, Polish, Russian, Swedish, Thai, Hindi, Italian, Korean, Bengali, Portuguese, Chinese, Hebrew, Serbian, Danish, Turkish, Greek, Indonesian, Zulu, Hungarian, Basque, Swahili, Afrikaans, Bosnian, Lao, Romanian, Slovenian, Ukrainian, Finnish, Malay, Javanese, Welsh, Bulgarian, Armenian, Icelandic, Vietnamese, Sinhalese, Maori, Gujarati, Kannada, Marathi, Tamil, Telugu, Amharic, Norwegian, Czech, Dutch. 
    
    \item Cipher Languages: Caesar1, Caesar2, Caesar3, Caesar4, Caesar5, Caesar6, Caesar7, Caesarneg1, Caesarneg2, Caesarneg3, Caesarneg4, Caesarneg5, Caesarneg6, Caesarneg7, Ascii, Hexadecimal, Base64, Leet, Vowel, Alphanumeric. A number in front of Caesar cipher means that the English alphabets were shifted by that much forward and a number in front of Caesarneg cipher means that the English alphabets were shifted by that much backward. 
    
\end{enumerate}

Out of these languages, we use the following for training our classifier:
Arabic, Chinese, Czech, Dutch, English, French, German, Hindi, Italian, Japanese, Korean, Polish, Portuguese, Russian, Spanish, Swedish, Thai, Bosnian, Turkish, Finnish, Indonesian, Bengali, Swahili, Vietnamese, Tamil, Telugu, Greek, Maori, Javanese, Caesar1, Caesar2, Caesar4, Caesarneg2, Caesarneg4, Caesarneg6, Ascii, Hexadecimal

And these for testing:
Persian, Croatian, Hebrew, Serbian, Danish, Zulu, Hungarian, Basque, Afrikaans, Lao, Romanian, Slovenian, Ukrainian, Malay, Welsh, Bulgarian, Armenian, Icelandic, Sinhalese, Gujarati, Kannada, Marathi, Amharic, Norwegian, Caesar, Caesar5, Caesar7, Caesarneg3, Caesarneg1, Caesar6, Caesarneg7, Caesarneg5, Base64, Alphanumeric, Vowel, LeetSpeak. 

%% file: tables/accuracy_by_resourceness.tex
\begin{tabular}{l|ccc}
\toprule
       & High-Res & Low-Res & Cipher \\
       \midrule
 LlamaGuard 1  & 69.92 & 41.25 & 16.07 \\
 LlamaGuard 2  & 75.28 & 62.2 & 16.2 \\
 LlamaGuard 3  & 82.23 & 75.84 & 24.74 \\
 AegisGuard (P) & 83.36 & 59.06 & 44.22 \\
 AegisGuard (D) & 88.14 & 76.26 & \textbf{83.21} \\
 WildGuard  & 81.35 & 43.51 & 16.53 \\
 HarmBench (llama) & 14.25 & 14.08 & 14.11 \\
 HarmBench (mistral) & 17.6 & 15.9 & 14.46 \\
 MD-Judge & 59.51 & 30.21 & 15.44 \\ 
 DuoGuard & 71.4 & 46.26 & 15.77 \\
 PolyGuard & \textbf{94.47} & 79.22 & 21.28 \\
 \toolname & 88.25 & \textbf{85.56} & 73.06 \\
 \bottomrule
\end{tabular}

%% file: appendix-folder/best-layer-experiment-details.tex
\section{Experimental Details of Filtering of Wikitext and its Translation}
\label{sec:wikitext-processing-details}

\begin{table}
    \centering
    \resizebox{\columnwidth}{!}{%
    \begin{tabular}{lcccc|c}
    \toprule
             & Thai Tox & Ukr Tox & TJS & ToxText & Avg. \\ 
    \midrule
    Layer 10   & 62.1 & 65.5 & 66.95 & 61.89 & 64.42 \\ 
    Layer 55   & 67.4 & 73.0 & \textbf{76.91} & 74.96 & 73.06 \\
    Layer 56   & 66.8 & 71.5 & 74.54 & 71.82 & 71.17 \\
    Selected layer 57   & \textbf{68.7} & 73.1 & 76.8 & \textbf{75.07} & \textbf{73.40} \\
    Layer 58   & 66.6 & 73.2 & 74.92 & 72.63 & 71.84 \\
    Layer 59   & 67.3 & \textbf{73.3} & 76.44 & 74.22 & 72.82 \\
    Layer 60   & 67.5 & 72.6 & 76.43 & 74.46 & 72.75 \\
    Layer 61   & 67.8 & 70.9 & 74.77 & 72.76 & 71.56 \\
    Layer 62   & 66.1 & 72.3 & 74.78 & 72.83 & 71.50 \\
    Layer 75   & 65.2 & 66.4 & 70.72 & 65.79 & 68.26 \\
    Last Layer & 63.1 & 51.2 & 61.33 & 56.76 & 59.05 \\ 
    \bottomrule
    \end{tabular}%
    }
    \caption{\toolname's accuracy of detecting harmful prompts when trained using representations from different model layers.}
    \label{tab:results-layer-ablation-merged}
\end{table}

%% file: appendix-folder/toolname-architecture.tex
\section{\toolname's Performance With Different Base LLMs}
\label{sec:base-llm-diff}

\begin{table*}
    \centering
    \resizebox{\textwidth}{!}{%
    \begin{tabular}{lcccccccc}
    \toprule
    Models & HarmBench & FQ & SimpleST & SaladB & TJS & ToxText & AdvBench & Avg. \\
    \midrule
    Llama3.1-8B-Instruct   & 81.16 & 88.72 & 88.78 & 87.35 & 71.32 & 69.20 & 88.62 & 82.16 \\
    Gemma-3-4B-Instruct    & 78.03 & 88.23 & \textbf{93.86} & 82.09 & 53.50 & 51.45 & 89.07 & 76.60 \\
    Qwen-3-4B-Instruct     & 82.06 & 86.25 & 80.99 & 82.15 & 58.83 & 57.70 & 84.80 & 76.11 \\
    Olmo-2-7B-Instruct     & 80.57 & 88.86 & 92.14 & \textbf{88.68} & 70.65 & 64.79 & 88.24 & 81.99 \\
    Mistral-8B-Instruct    & 84.27 & 87.23 & 89.82 & 88.08 & 72.49 & 68.71 & \textbf{89.07} & 82.81 \\
    Llama3.3-70B-Instruct  & \textbf{89.13} & \textbf{89.57} & 89.62 & 87.30 & \textbf{76.68} & \textbf{75.07} & 86.59 & \textbf{84.85} \\
    \bottomrule
    \end{tabular}%
    }
    \caption{\toolname's accuracy of detecting harmful prompts when paired with different underlying LLMs across multiple benchmarks. }
    \label{tab:results-llm-comparison}
\end{table*}